\documentclass[letterpaper, 10 pt, conference]{ieeeconf}  % Comment this line out if 
\IEEEoverridecommandlockouts                              % This command is only needed if 
\usepackage{graphics} % for pdf, bitmapped graphics files
\usepackage{epsfig} % for postscript graphics files
\usepackage{mathptmx} % assumes new font selection scheme installed
\usepackage{times} % assumes new font selection scheme installed
\usepackage{amsmath} % assumes amsmath package installed
\usepackage{amssymb}  % assumes amsmath package installed
\usepackage{float}
\usepackage{array}
\usepackage{mdwmath}
\usepackage{mdwtab}
\usepackage{todonotes}
\usepackage[switch, pagewise]{lineno}
\usepackage{caption}
\usepackage{subcaption}
\usepackage{subfiles}
\usepackage[nolist]{acronym}
\usepackage[version-1-compatibility]{siunitx}
\usepackage{calrsfs}
\usepackage{setspace}
\usepackage{multirow}
\usepackage{bm}
\usepackage[mathscr]{euscript}
\usepackage{pdfpages}

\newcommand{\etal}{~et al.~}
\newcommand{\Rnum}{\mathbb{R}} % Symbol fo the real numbers set

\DeclareMathAlphabet{\pazocal}{OMS}{zplm}{m}{n}
\renewcommand*{\arraystretch}{.5}
\makeatletter
\renewcommand*\env@matrix[1][\arraystretch]{%
	\edef\arraystretch{#1}%
	\hskip -\arraycolsep
	\let\@ifnextchar\new@ifnextchar
	\array{*\c@MaxMatrixCols c}}
\makeatother
\newcounter{inlineenum}
\renewcommand{\theinlineenum}{\arabic{inlineenum}}
\newenvironment{inlineenum}
{\unskip\ignorespaces\setcounter{inlineenum}{0}%
	\renewcommand{\item}{\refstepcounter{inlineenum}{\textit{\theinlineenum})~}}}
{\ignorespacesafterend}

\newcommand{\change}[1]{\textcolor{black}{#1}}
\newcommand{\changerev}[1]{\textcolor{black}{#1}}

\graphicspath{ {./} } 
\begin{acronym}[RCF] % Give the longest label here so that the list is nicely 
\acro{RCF}{Reactive Controller Framework}
\acro{CNN}{Convolutional Neural Network}
\acro{MPC}{Model Predictive Control}
\acro{LF}{left front}
\acro{RF}{right front}
\acro{LH}{left hind}
\acro{RH}{right hind}
\acro{IMU}{Inertial Measurement Unit}
\acro{LO}{Leg Odometry}
\acro{EKF}{Extended Kalman Filter}
\acro{DDT}{Drift per Distance Traveled}
\acro{AICP}{Auto-tunned Iterative Closest Point}
\acro{FF}{feedforward}
\acro{GRFs}{ground reaction forces}
\acro{GRF}{ground reaction force}
\acro{COM}{Center of Mass}
\acro{CTT}{\ac{COM} tracking task}
\acro{CST}{contact sequence task}
\acro{VFA}{Vision-based Foothold Adaptation}
\acro{LF}{left-front}
\acro{RF}{right-front}
\acro{LH}{left-hind}
\acro{RH}{right-hind}
\acro{HPU}{Hydraulic Pump Unit}
\acro{HPUs}{Hydraulic Pump Units}
\acro{DOFs}{degrees of freedom}
\acro{DOF}{degree of freedom}
\acro{ISA}{Integrated Smart Actuator}
\acro{IIT}{Istituto Italiano di Tecnologia}
\acro{VO}{Visual Odometry}
\acro{REU}{Remote Electronics Unit}
\acro{LIP}{Linear Inverted Pendulum}
\acro{QP}{Quadratic Program}
\acro{NMPC}{Nonlinear Model Predictive Control}
\acro{iLQR}{Iterative Linear Quadratic Regulator}
\acro{NLP}{Nonlinear Program}
\acro{MICP}{Mixed-Integer Convex Program}
\acro{ZOH}{zero-order hold}
\acro{JSIM}{joint-space inertia matrix}
\acro{RMS}{root mean square}
\end{acronym}
\begin{document}
	\title{\LARGE \bf
			MPC-based Controller with Terrain Insight for Dynamic Legged Locomotion 
		  }	
	
	\author{Octavio Villarreal$^1$, Victor Barasuol$^1$, Patrick \changerev{M.} 
	Wensing$^2$, 
	\change{	Darwin 
		G. 
		Caldwell$^3$}, and Claudio Semini$^1$% <-this %%%stops a space
			\thanks{$^1$Dynamic Legged Systems lab, Istituto Italiano di 
			Tecnologia, Via Morego 30, 16163 Genoa, Italy.
			{\tt\small firstname.lastname@iit.it}}%
			\thanks{$^2$ Department of Aerospace and Mechanical Engineering, 
			University of Notre Dame, IN 46556 USA.
			{\tt\small pwensing@nd.edu}}
			\thanks{$^3$ Department of Advanced 
			Robotics, Istituto Italiano di Tecnologia, Via Morego 30, 16163 
			Genoa, Italy.
			{\tt\small darwin.caldwell@iit.it}}%
		   }
		   \null
		   \includepdf[pages=-]{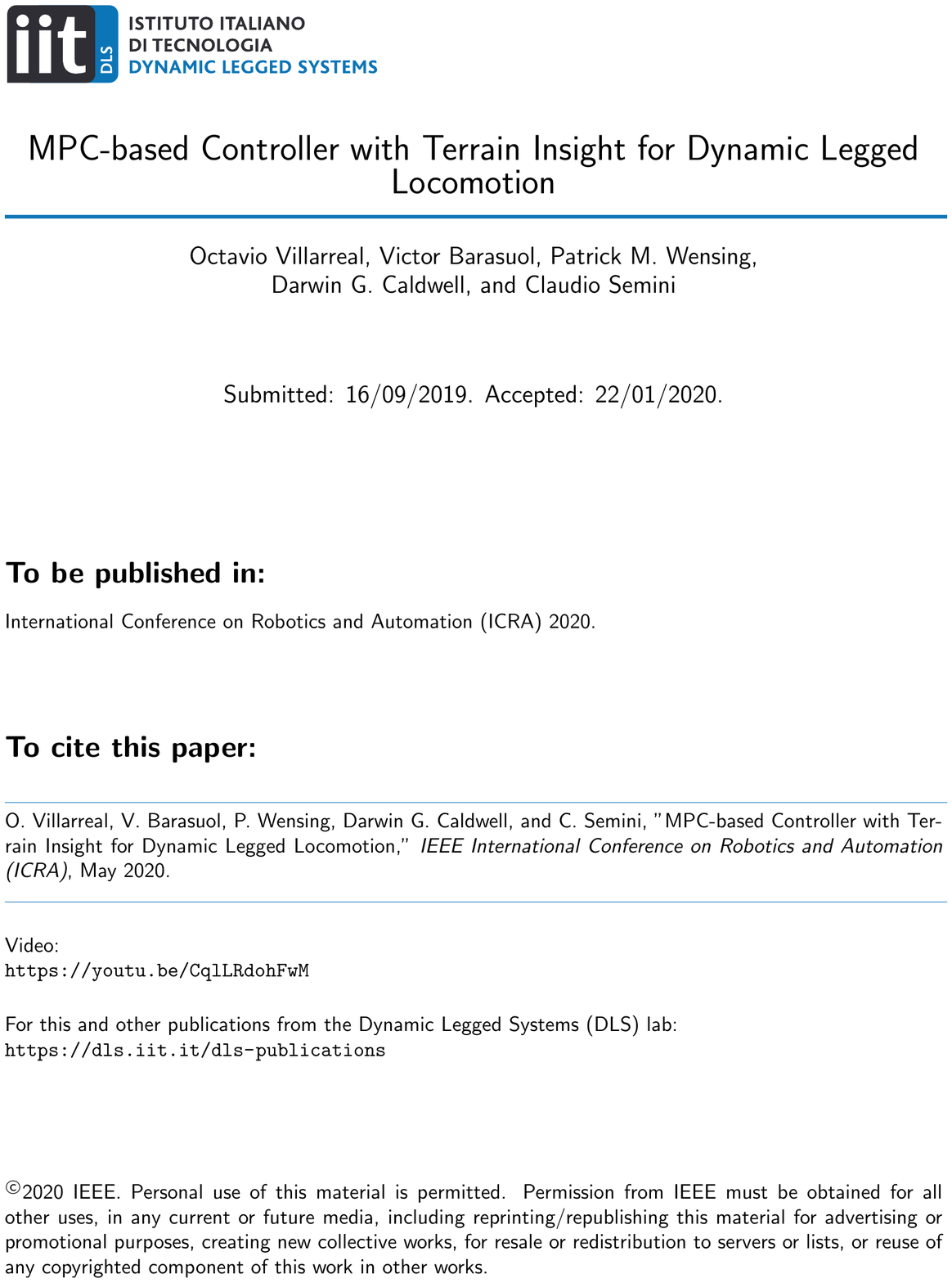}
		
	\maketitle
	\thispagestyle{empty}
	\pagestyle{empty}
%		
	%%%%%%%%%%%%%%%%%%%%%%%%%%%%%%%%%%%%%%%%%%%%%%%%%%%%%%%%%%%%%%%%%%%%%%%%%%%%
	\begin{abstract}
		We present a novel control strategy for dynamic legged locomotion in 
		complex scenarios that considers information about the morphology of 
		the terrain in contexts when only on-board mapping and computation are 
		available. The strategy is built on top of two main elements: first a 
		contact sequence task that provides safe foothold locations based on a 
		convolutional neural network to perform fast and continuous evaluation 
		of the terrain in search of safe foothold locations; then a model 
		predictive controller that considers the foothold locations given by
		the contact sequence task to optimize target ground reaction forces.
		We assess the performance of our strategy through simulations of the hydraulically actuated quadruped robot \textit{HyQReal} traversing rough terrain under realistic on-board sensing and computing conditions.
	\end{abstract}	
	%%%%%%%%%%%%%%%%%%%%%%%%%%%%%%%%%%%%%%%%%%%%%%%%%%%%%%%%%%%%%%%%%%%%%%%%%%%%
	\section{Introduction}	
	\label{section:introduction}
	 
	Considering terrain morphology 
	allows legged robots to traverse 
	more complex scenarios (e.g., \cite{kalakrishnan09iros,fankhauser18icra,belter11jfr}). 
	Nevertheless, building a model of the terrain is often computationally costly, mainly because 
	of the dense nature of visual data. On top of the mapping problem, 
	\change{feasible contact sequences are needed} to traverse the 
	terrain safely. Computing these 
	contact sequences can also be costly \cite{lin19icra,tonneau18tro}. In general, strategies that 
	consider 
	visual information of the terrain are mostly focused on trajectory optimization 
	\cite{farshidian17icra,winkler18ral,fernbach18iros,herzog15humanoids}. In most approaches, 
	contact 
	sequences and \changerev{the} \ac{COM} trajectory are computed prior to the 
	motion, or 
	are limited to (quasi-) statically stable gaits to not compromise stability 
	due to time 
	constraints \cite{fankhauser18icra,kalakrishnan2011,belter19icra}. 
	
	In this work, we combine the low computational time from our previous \ac{VFA} strategy from \cite{villarreal19ral} with a 
	\ac{MPC}-based trunk controller. These two approaches are mutually beneficial to each 
	other. On one hand, we 
	exploit the computational gain that we obtain from the \ac{CNN} in the \ac{VFA}
	strategy, to generate safe contact sequences to be used in the \ac{MPC}-based \ac{COM} tracking 
	controller. 
	On the other hand, the \ac{VFA} benefits from the \ac{MPC}-based controller with respect to the foothold prediction. A \textit{foothold prediction} is a future landing position based on the nominal trajectory of the legs and the trunk velocity. We stress that the foothold predictions are different from the state predictions computed using the \ac{MPC}. 
	
	We start from the premise that optimizing \acp{GRF} accounting for future 
	states using \ac{MPC} will lead 
	to 
	better foothold predictions, since they depend both on foot trajectories and robot states.
	If the robot states have large acceleration peaks, the foothold prediction is affected negatively.
	An improved selection of the desired \acp{GRF} would reduce acceleration 
	peaks, improving foothold predictions. 
	This allows the 
	robot to handle more 
	difficult scenarios, such as changes in elevation and orientation, in a safer and more reliable
	way, as demonstrated in simulations.
	
	We perform simulations using the quadruped robot \textit{HyQReal} \cite{hyqreal19irim}. Its four legs weigh 
	in total \SI{48}{\kilogram} (between 37\% and 45\% of the total weight of the robot, with and 
	without on-board hydraulic/electric power units, respectively). 
	This means that when fast 
	motions are required, swing legs play a significant role in the robot dynamics.
	In this paper, we directly 
	compensate for these effects by computing the wrench on the body due to the desired joint accelerations of 
	the legs,
	improving state tracking and foothold prediction. 	
	
	The result is a stable locomotion strategy, which is robust to a wide range 
	of 
	disturbances and is able to act preemptively to obstacles based on visual information.
	\begin{figure*}[h!]
		\centering
		\begin{subfigure}[t]{0.74\linewidth}
			\centering
			\includegraphics[width=1\linewidth]{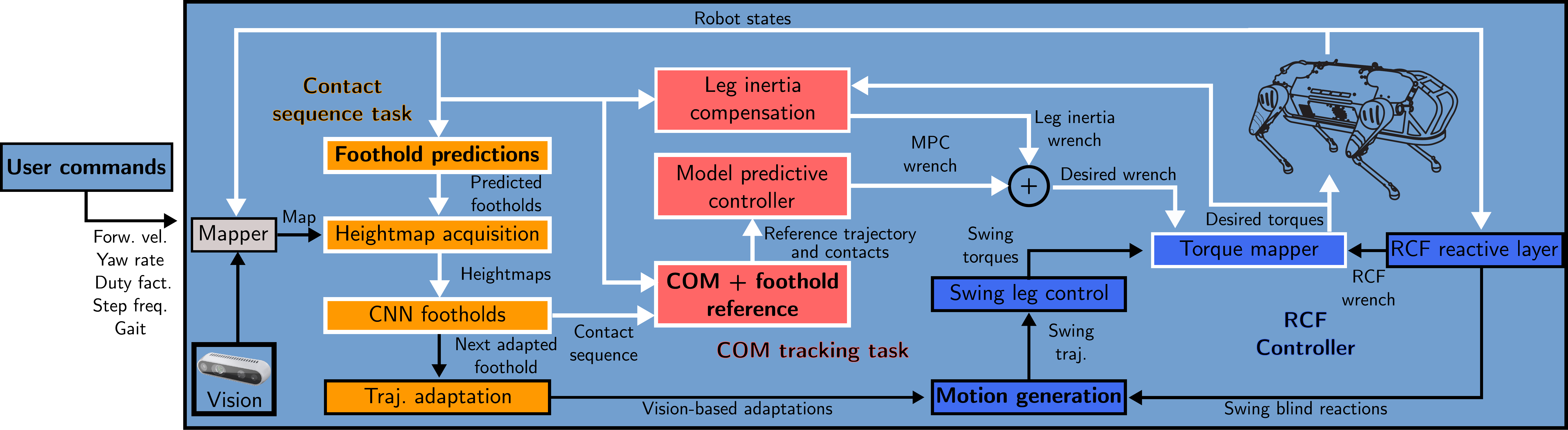}
		\end{subfigure}
		\begin{subfigure}[t]{0.24\linewidth}
			\centering
			\includegraphics[width=0.9\linewidth]{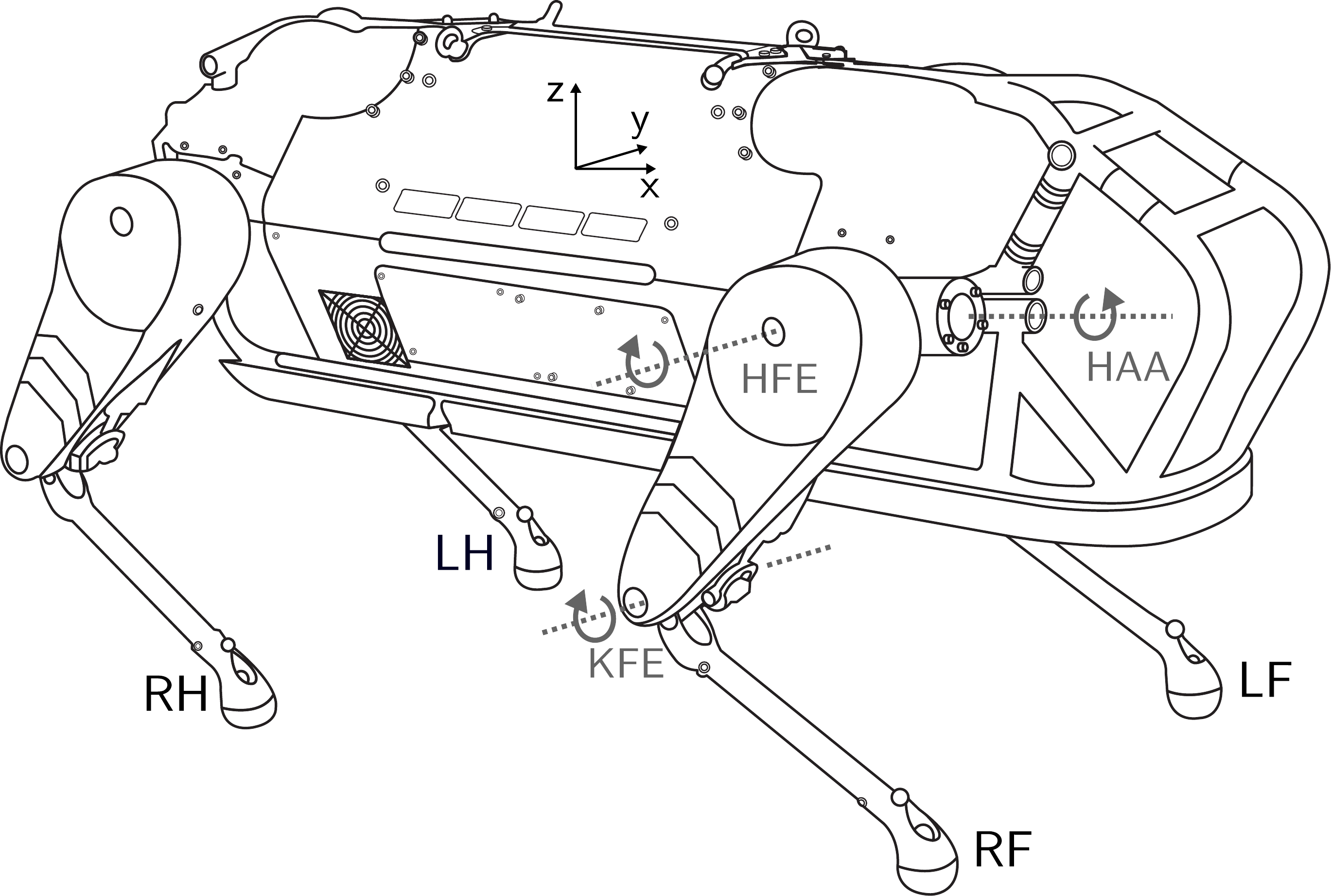}		
		\end{subfigure}
		\caption{\small \changerev{L}eft: schematic drawing describing 
		\changerev{the} 
		control
		strategy. The user commands are used by the \textit{foothold 
		predictions}, \textit{COM + reference}, and \textit{motion generation} 
		blocks (denoted with bold text). The contact sequence task is denoted 
		by the orange blocks. The COM tracking task is denoted by the red 
		blocks. The RCF \cite{barasuol13icra} (blue blocks) serves as an 
		interface for the VFA, the MPC + leg inertia compensation controller 
		and a reactive layer for \changerev{``}blind'' locomotion. \change{The 
		torque 
		mapper 
		block distributes the total wrench among the \acp{GRF} (solving a QP 
		similarly to 
		\cite{focchi2016}) 
		and maps to joint torques using $\bm{\tau} = \mathbf{J}^\intercal 
		\mathbf{F}$.} \change{The white arrows 
		connect the blocks where the MPC and the CNN interact.} Right: joint 
		and leg definitions of HyQReal.}
		\label{fig:strategy}
		\vspace{-0.42cm}
	\end{figure*}
	We summarize the contributions of this paper as follows:
	\begin{itemize}
		\item We devised a locomotion strategy that evaluates the terrain, generates safe contact 
		sequences and allows for dynamic locomotion in difficult 
		scenarios. The new strategy displayed an improvement in foothold 
		prediction with respect to \cite{villarreal19ral}, reducing the 
		prediction error by a percentage between 7\% to 36\%. 
%		(approximately between \SI{0.6}{\centi \meter} and \SI{3}{\centi 
%\meter}).
		\item We combine a \ac{CNN}-based foothold adaptation 
		strategy and an \ac{MPC}-based trunk controller and show how they 
		mutually benefit from each other. To the best of our 
		knowledge, this is the first time that an \ac{MPC}-based locomotion 
		controller uses the terrain geometry in combination with a machine 
		learning strategy.
%		\item \change{We provide empirical proof of the improvement in 
%		performance of the \ac{MPC}-based controller when compensating  for 
%		the wrench exerted by the legs during swing phase due to their inertia. 
%		The compensation renders the model used for prediction in the 
%		\ac{MPC} more representative, since 
%		the leg inertia is handled separately, further reducing the error in 
%		foothold prediction.}
		\item We improve the performance of the \ac{MPC}, by 
		compensating for the wrench exerted by the legs during swing 
		phase due \changerev{to the large weight of the legs with respect to 
		the 
		total weight of the robot}. This compensation renders the model 
		used for prediction in 
		the \ac{MPC} more representative, 
		since the leg inertia is handled separately, further reducing the error 
		in foothold prediction.
	\end{itemize}
	This paper is structured as follows: Section \ref{section:related_work} summarizes 
	the previous work relevant to this research; 
	Section 
	\ref{section:locomotion_strategy} details the methodology used to derive our locomotion 
	strategy; Section \ref{section:results} summarizes our results. 
	Conclusions and future work are presented in Section \ref{section:conclusions}. 
	\section{Related Work} 
	\label{section:related_work}
	
	The spectrum of strategies when using \ac{MPC} varies mostly depending on the trade-off between 
	model accuracy and 
	computational cost.
	The work of Di Carlo\etal 
	\cite{iros18dicarlo} considers a simplified version of the centroidal dynamics 
	model, neglecting leg inertia, and ignoring the effects of non-zero 
	roll/pitch on the dynamics of the body. The 
	optimization problem 
	is still convex and it is solved in real-time as a \ac{QP}. The strategy keeps the 
	robot in balance during a range of highly 
	dynamic 
	motions (e.g., trot, bound, and gallop) on the Cheetah 3 robot 
	\cite{bledt18iros}. \change{Herein, a feedforward torque compensates leg 
	inertia to improve swing leg trajectory tracking. Our work differs 
	from this implementation, since we compute the wrench exerted on the trunk 
	by the swing legs and compensate for it using the \acp{GRF} of the stance 
	legs to improve \ac{COM} tracking.}
	
	Some other approaches tackle the trade-off between computational cost and 
	model accuracy 
%	not by simplifying the whole-body dynamics, but 
	by relying on reducing the computational cost of the optimization problem 
	solver. A 
	remarkable example is the one devised by Neunert\etal in 
	\cite{neunert17ral2}. Their approach performs \ac{NMPC} and 
	relies on a custom solver based on the \ac{iLQR} algorithm and exploits automatic 
	differentiation \cite{giftthaler17ar}. 
	
	Using \ac{MPC} has provided a systematic and robust way to 
	\changerev{address} the quadruped 
	locomotion problem. Nevertheless, in general it does not take into account 
	future terrain within \changerev{the} prediction. Instead, it reacts to the 
	terrain and relies on the fact that the continuous update of the state for 
	the 
	initialization of the optimization provides enough robustness. It is still challenging to leverage terrain information to improve 
	the performance of 
	\ac{MPC} strategies.
	
	There are some trajectory optimization methods that consider terrain information for 
	quadrupeds. The method devised by Winkler\etal in \cite{winkler18ral} is able to optimize gait, 
	\ac{COM} 
	trajectory and contacts on non-flat terrain based on a simplified centroidal dynamics model 
	using an off-the-shelf \ac{NLP} solver. In a similar fashion, 
	Aceituno\etal showcase 
	a motion 
	planning algorithm that computes gait pattern, contact sequences and \ac{COM} 
	trajectory as an 
	outcome of a \ac{MICP} on several non-planar convex surfaces in 
	\cite{aceituno2017ral}. 
	However, in their work\changerev{,} either the trajectory is computed only 
	once before 
	\change{execution}, or 
	the terrain is assumed to be known and there are no 
	experiments with 
	vision sensors during \change{the motion}.  
	
	The early works of Kalakrishnan\etal \cite{kalakrishnan09iros,kalakrishnan2011} on 
	LittleDog 
	pioneered methods to include vision sensors for locomotion 
	by relying on external motion capture systems. Belter\etal \cite{belter11jfr} and 
	Fankhauser\etal 
	\cite{fankhauser18icra} devised control architectures that allowed their legged 
	platforms to 
	traverse complex scenarios only using on-board sensing enhanced with vision, which 
	were mostly 
	demonstrated for statically stable motions. \changerev{In} our previous 
	work \cite{villarreal19ral}, we 
	presented a 
	strategy that was able to adapt footholds based on a \ac{CNN}. The approach generated swing leg 
	trajectory adaptations in less than \SI{0.1}{\milli \second}. This 
	allowed us to execute dynamic locomotion in complex scenarios. 

	\section{Locomotion Strategy} 
	\label{section:locomotion_strategy}
	
%	\vspace{-0.15cm}
	Our goal is to produce robust and stable locomotion in complex scenarios using terrain 
	information provided by on-board vision sensors. We combine \ac{MPC} with a \ac{CNN}-based foothold 
	adaptation strategy \cite{villarreal19ral}. The combination of these two strategies benefits each other. 
	
	State predictions in the \ac{MPC} are computed using the centroidal dynamics model and safe contact sequences are based on the \ac{VFA}. 
	Future footholds can be continuously computed by \change{the} \ac{VFA} 
	approximately every \SI{0.5}{\milli \second},
	enabling the \ac{MPC} to reason about the effects of future contacts, 
	without having to consider them \changerev{as optimization variables}. We 
	then build upon these 
	contacts to provide the reference pose for the robot along the prediction horizon.
	
	The block diagram shown in Fig.~\ref{fig:strategy} describes our locomotion 
	strategy. It 
	entails 
	three main elements: the contact sequence task, the \ac{COM} tracking task and the \ac{RCF} \cite{barasuol13icra}. The contact sequence task provides the future contact locations according to the 
	robot current states and the gait timing parameters. The \ac{COM} tracking task is in charge 
	of both 
	generating and following a \ac{COM} trajectory according to the contact sequence task, the 
	\changerev{current robot} states, and the gait parameters. We use the 
	\ac{RCF} \cite{barasuol13icra} as 
	controller interface. This modular 
	framework allows us to combine the \textit{RCF reactive layer} block in Fig. \ref{fig:strategy} with our vision-based strategy. This layer is comprised by several modules that allow the robot to perform robust locomotion in rough terrain only using proprioception. 
	\change{The \ac{RCF}
	combines these reactive modules with the \ac{MPC} and the \ac{VFA}.}  
	
	The user 
	commands 
%	(see Fig. \ref{fig:strategy}) 
	are: forward velocity 
	$\mathbf{V}_f\in\Rnum^2$ 
	($x$ and $y$ velocities), 
	yaw rate $\dot{\psi}_{ref}\in\Rnum$, duty factor $D_f$, step frequency 
	$f_s$ and gait 
	$\pazocal{G}$. 
	$\mathbf{V}_f$ and $\dot{\psi}_{ref}$ are provided via joystick commands.
%	\changerev{The rest are user-defined parameters according to the desired 
%	gait and speed.}
	The rest of the 
	parameters are 
	preset by the user according to the desired gait and range of speeds.
	
	Below we explain the two main elements of our strategy: 
	the contact sequence and the \ac{COM} tracking tasks.
	\subsection{Contact Sequence Task}
	\label{section:contact_task}
	We extend the \changerev{use} of the \ac{VFA} \cite{villarreal19ral} to 
	provide the 
	subsequent \changerev{eight} 
	reference footholds (two strides). These footholds are used to generate the \ac{COM} 
	reference trajectory and provide the contacts \change{for} the model 
	described 
	in Section~\ref{section:com_task}.
	
	\paragraph{Vision-based Foothold Adaptation}
	\changerev{T}he purpose of the \ac{VFA} is to continuously compute 
	adjustments for the 
	trajectory of the feet in order to 
	avoid collisions and unsafe or unreachable landing positions. For a more 
	detailed description on 
	this method we refer the reader to \cite{villarreal19ral}.
	 
	For a leg in swing phase, we initially compute a prediction of its landing position based on 
	the current velocity of the trunk (taken from the state estimator) and the trajectory of the 
	foot (in our case a half-ellipse) using the approximation:
	\begin{equation}
		\hat{\mathbf{p}}_{i} = \bar{\mathbf{p}}_{i} + 
		\frac{1}{2}\change{\bm{\ell}}_s + \Delta t_i 
		\dot{\mathbf{r}}
		\label{eq:foothold_prediction}
	\end{equation}
	where $\hat{\mathbf{p}}_i\in\Rnum^3$ is the \textit{predicted foothold} of 
	\changerev{$i$-th leg, for $i = 1,...,l$, with $l$ \changerev{being the 
	total number of legs} (see Fig.~\ref{fig:strategy})},
%	 $i\changerev{\in \mathscr{L}}$ for 
%	$\mathscr{L} = 
%	LF,RF,LH,RH$ \changerev{(see Fig.~\ref{fig:strategy})}, 
	$\bar{\mathbf{p}}_i\in\Rnum^3$ 
	is the 
	center of the ellipse,
%	 of leg $i$, 
	 $\Delta t_i$ is 
	the time remaining 
	to the next stance 
	change,
%	 of leg $i$, 
	$\change{\bm{\ell}}_s~\in~\Rnum^3$ is the step length 
	vector, and $\dot{\mathbf{r}}\in \Rnum^3$ corresponds to 
	the velocity of the base. All vector variables are given in world coordinates. In the case of 
	the next touchdown of a swing leg, $\Delta t_i = \frac{1 - D_f}{f_s} - t_{sw,i}$, where $D_f$ 
	is 
	the duty factor, $f_s$ is the step frequency and $t_{sw,i}$ is the elapsed swing time since the 
	latest
	lift-off.
%	 of leg $i$. 
	 The first two terms in \eqref{eq:foothold_prediction} are related to the 
	leg trajectory, while the third term is related to the displacement of the base.
	
	After computing the prediction of the next foothold, a 2D representation of the terrain around 
	that foothold is acquired, namely a \textit{heightmap}.
	We pre-train a \ac{CNN} to learn the \textit{optimal} 
	footholds from 
	heightmaps \cite{villarreal19ral} \change{considering collisions, 
	terrain roughness\changerev{,} and process uncertainty. The architecture of 
	the \ac{CNN} 
	is designed as a trade-off between prediction accuracy and speed}. The 
	\ac{CNN} takes on average \SI{0.1}{\milli \second} to evaluate a 
	heightmap and 
	output a safe foothold. This fast computation time allows us to continuously adapt the 
	trajectory of the swing leg to reach the adapted foothold.	
	\paragraph{Reference Contact Sequence} 
	We use the computational gain obtained by the \ac{VFA} to 
	evaluate further ahead in the terrain. 
	Knowing that the gait is periodic and defined by the step frequency $f_s$ and the duty factor $D_f$, 
	we can estimate the timings 
	for the non-immediate foot contacts. Using these timings, one can compute the predicted 
	foothold locations for each of the legs at every stance change (lift-off or touchdown)	replacing them for $\Delta t$ in \eqref{eq:foothold_prediction}. We then use our \ac{CNN}-based 
	foothold adaptation to adjust the predicted foothold location. This 
	\changerev{calculation} is done for the next two 
	gait 
	cycles (eight contacts in total and 16 stance changes). An 
	example of a safe foothold sequence can be seen \changerev{on the right 
	side of the series of snapshots of Fig. \ref{fig:performance_task_sim}}. 
	Namely, $\mathbf{p}_i[\changerev{\kappa}]$ is the 
	contact 
	location of leg $i$ at stance change \changerev{$\kappa$, for $\kappa = 
	0,...,16$}. In \eqref{eq:foothold_prediction}, $\dot{\mathbf{r}}$ is assumed constant in between stance changes.
	
	The \ac{CNN} continuously provides safe contact sequences at task frequency 
	(\SI{250}{\hertz}). These sequences are used both as future foot positions 
	and to inform the \ac{MPC} controller to improve the \ac{COM} regulation, 
	as explained in Section \ref{section:com_task}. This 
	interaction is \change{shown} in Fig. \ref{fig:strategy}. One key feature 
	of 
	\change{the} approach is that safe footholds are computed without including 
	them as optimization variables in the \ac{MPC} controller, which 
	significantly decreases the complexity of the problem.
	\subsection{\ac{COM} Tracking Task}
	\label{section:com_task}
	\paragraph{\ac{COM} Reference Generation} To provide the reference trajectory for the \ac{COM} 
	along the prediction horizon, we compute 
	its location at every stance change based on the desired gait timings using $f_s$ and $D_f$. For two 
	gait cycles, there are a total of 16 stance changes, so we compute a total of 16 \ac{COM} 
	positions.
	 Similarly to the third term of \eqref{eq:foothold_prediction}, we compute 
	the reference yaw using the desired yaw rate as
	\begin{equation}
	\psi_{ref}[\change{\kappa}] = \psi + \Delta t[\change{\kappa}] 
	\dot{\psi}_{ref}
	\end{equation}
	where $\psi_{ref}[\change{\kappa}] \in \Rnum$ is the yaw reference at 
	stance change 
	\change{$\kappa$}, for \changerev{$\kappa = 0,...,16$,} $\psi \in \Rnum$ is 
	the current yaw, and
%	\changerev{Herein, $\Delta t[\kappa] = \frac{1-D_f}{f_s} - 
%	t_{sw}[\kappa]$, 
%	where $t_{sw}[\kappa]$ is the elapsed swing time from the latest stance 
%	change $\kappa$.}
	\changerev{$\Delta t[\kappa]$ is the time for a stance change to 
	happen from $\kappa = 0$.}
%	angle of the body, 
%	, and $\Delta t[\change{\kappa}]$ is the remaining time 
%	before the 
%	next stance change \change{$\kappa + 1$}. 
	Using the reference for the yaw 
	angle, we compute the reference 
	position of the \ac{COM} with respect to the world 
	\begin{equation}
		\mathbf{r}_{ref}[\change{\kappa}] = \mathbf{r}  + \Delta 
		t[\change{\kappa}]\mathbf{R}_z(\Delta 
		\psi\changerev{[\kappa]})\dot{\mathbf{r}}_{ref}  
	\end{equation}
	where $\mathbf{r}_{ref}[\change{\kappa}]\in\Rnum^3$ is the reference 
	position for the \ac{COM} at stance 
	change $\change{\kappa}$, 
	$\mathbf{R}_z(\Delta \psi) \in \Rnum^{3x3}$ is the rotation matrix around the $z$ axis about 
	$\Delta \psi\changerev{[\kappa]}$ (with $\Delta 
	\psi\changerev{[\kappa]} = \psi_{ref}[\change{\kappa}] - \psi$) 
	and 
	$\dot{\mathbf{r}}_{ref}$ is the reference velocity obtained from $\mathbf{V}_f$ and 
	$\dot{\psi}_{ref}$. This provides the reference for the next $x$ and $y$ positions of the 
	\ac{COM} with respect to the world. 
	
	The reference for the body roll $\phi_{ref}$ and pitch $\theta_{ref}$ relies on the contact configuration 
	at each stance change. We 
	estimate the orientation of the terrain and define that orientation as reference for the body. 
	We also use the contacts to define a height $z$ reference position for the body (namely, 
	$r_{ref,z}[\change{\kappa}]$), setting it to remain at a constant distance 
	from the 
	center position of the approximated plane in the direction of the $z$ world 
	axis. To obtain 
	$\dot{r}_{ref,z}[\change{\kappa}]$, 
	$\dot{\phi}_{ref}[\change{\kappa}]$ and 
	$\dot{\theta}_{ref}[\change{\kappa}]$ we derive 
	numerically 
	between 
	samples of $r_{ref,z}[\change{\kappa}]$, $\phi_{ref}[\change{\kappa}]$ and 
	$\theta_{ref}[\change{\kappa}]$, respectively. Finally, we 
	evenly 
	sample the 16 reference points given by the stance changes, filling the gaps in between samples 
	using a \ac{ZOH}. We define a reference vector at evenly sampled time $k$ as
	\begin{equation}
		\mathbf{x}_{ref}[k] = \begin{bmatrix}
		\Theta_{ref}^{\top}[k] & 
		\mathbf{r}_{ref}^{\top}[k] 
		&\dot{\Theta}_{ref}^{\top}[k] & 
		\dot{\mathbf{r}}_{ref}^{\top}[k]
		\end{bmatrix}^\top
	\end{equation}
	with $\Theta_{ref}[k] = \left[\theta_{ref}[k]\text{ }\phi_{ref}[k]\text{ } 
	\psi_{ref}[k]\right]^\top$ and 
	$\mathbf{r}_{ref}[k] = \left[ r_{ref,x}[k]\text{ }r_{ref,y}[k]\text{ 
	}r_{ref,z}[k]\right]^\top$. 
	Figure~\ref{fig:performance_task_sim} shows multiple \ac{COM} references.
%	A series of \ac{COM} references is shown in Fig. 
%	\ref{fig:performance_task_sim}. 
	\paragraph{Dynamic Model} \changerev{O}ur \ac{MPC} trunk balance controller 
	is inspired by the 
	work of Di Carlo\etal \cite{iros18dicarlo}. We model the robot as a rigid 
	body subject to 
	contact patches at each stance foot and we neglect the effects of 
precession and nutation as in 
	\cite{focchi2016}. 
	However, 
	there are two key 
	differences in our approach: 
	firstly, we do not define the reference roll and pitch angles to be zero. Additionally, although we 
	do not explicitly consider the leg inertia in 
	our model for control, we compensate for it by computing the wrench exerted by the legs 
	using the actuated part of the joint-space inertia matrix and the desired accelerations of the 
	joints. We explain how this is done by the end of this section.
	
	The dynamics of the rigid body 
	and its rotational kinematics are given by
	\begin{align}
	\ddot{\mathbf{r}} &= \frac{\sum_{i = 1}^{\changerev{l}} \mathbf{F}_i}{m} + 
	\mathbf{g}  
	\label{eq:linear_acc} \\ 
	\mathbf{I}\dot{\mathbf{\omega}} &= \sum_{i=1}^{\changerev{l}} 
	\mathbf{p}_i 
	\times \mathbf{F}_i 
	\label{eq:angular_acc} \\ 
	\dot{\mathbf{R}} &= [\omega]_\times \mathbf{R}
	\label{eq:rot_kin}
	\end{align}
	where $\mathbf{r}\in \Rnum^3$ is the position of the \ac{COM}, $\mathbf{F}_i\in \Rnum^3$ is the 
	\ac{GRF} at foot
	$i$, $m\in\Rnum$ is the robot mass, $\mathbf{g}\in\Rnum^3$ is the gravitational acceleration, 
	$\mathbf{I}\in\Rnum^{3\times 3}$ is the inertia tensor of the robot, $\mathbf{p}_i\in\Rnum^3$ 
	is the 
	$i$-th foot contact position, $\mathbf{R}\in\Rnum^{3\times 3}$ is the rotation matrix from body to world coordinates according to roll $\phi$, pitch $\theta$ and yaw 
	$\psi$ angles and $\mathbf{\omega}\in\Rnum^3$ is the robot's angular velocity. The operator 
	$[\mathbf{x}]_\times$ is the skew-symmetric matrix such that $[\mathbf{x}]_\times \mathbf{y} = 
	\mathbf{x} 
	\times \mathbf{y}$.	In \eqref{eq:angular_acc} we are neglecting precession and nutation 
	effects, namely 
	$\mathbf{\omega}\times 
	\mathbf{I}\mathbf{\omega}\approx 0$.
	We rewrite equations 
	\eqref{eq:linear_acc}, 
	\eqref{eq:angular_acc} and \eqref{eq:rot_kin} in state-space representation. Initially, from 
	\eqref{eq:rot_kin} we can obtain the angular velocity in terms of the body's Euler 
	angles \changerev{from}
	\begin{equation}
		[\mathbf{\omega}]_\times =  \dot{\mathbf{R}}\mathbf{R}^\top
	\end{equation}
	which can be rewritten as
	\begin{equation}
		\mathbf{\omega} = \mathbf{T}(\Theta)\dot{\Theta}
		\label{eq:angular_velocity}
	\end{equation}
	where $\Theta = [\phi\text{ }\theta\text{ }\psi]^\top$ and 
	$\mathbf{T}(\Theta)$ is the matrix that maps from \changerev{E}uler angle 
	rates to 
	angular velocities. The only condition on $\mathbf{T}(\Theta)$ to be 
	invertible is $\theta\ne \pi/2$, which in practice does not happen (it 
	implies that the robot is pointed vertically). Thus, the angular rate can 
	be obtained as
	\begin{equation}
	\dot{\Theta} = \mathbf{T}^{-1} (\Theta) \mathbf{\omega}
	\label{eq:angular_rates}
	\end{equation} 
	
	We define state vector\change{\footnote{\change{Herein, 
				$\mathbf{g}$ is appended in the state vector to reach the form 
				given in 
				\eqref{eq:compact_dynamics}}}} $\mathbf{x} 
				=[\Theta\change{^\top}\text{ 
	}\mathbf{r}\change{^\top}\text{ 
	}\mathbf{\omega}\change{^\top}\text{ 
	}\dot{\mathbf{r}}\change{^\top}\text{ 
	}\mathbf{g}\change{^\top}]^\top$ \change{and rearrange} 
	\eqref{eq:linear_acc}, 
	\eqref{eq:angular_acc} and 
	\eqref{eq:angular_rates} \changerev{to write them in state-space as}
	\begin{equation}
	\dot{\mathbf{x}}(t) = \mathbf{A}(\Theta)\mathbf{x}(t) + 
	\mathbf{B}(\Theta,\mathbf{p}_{LF},...,\mathbf{p}_{RH})\mathbf{u}(t) 
	\label{eq:compact_dynamics}
	\end{equation}
	where $\mathbf{u}$ is the vector of \acp{GRF}.  Note that no assumptions 
	are made about the 
	orientation 
	of the robot\footnote{If $\theta \approxeq \phi \approxeq 0$ then: $\dot{\mathbf{x}}(t) = \mathbf{A}(\psi)\mathbf{x}(t) + 
		\mathbf{B}(\psi,\mathbf{p}_{LF},...,\mathbf{p}_{RH})\mathbf{u}(t) 
		\label{eq:compact_dynamics_yaw}$} (except for $\theta \ne 
		\changerev{\pi/2}$) and we 
		explicitly denote the dependence of 
	$\mathbf{T}$ and $\mathbf{I}$ \change{on} $\Theta$. 
	
	In a similar fashion to \cite{iros18dicarlo}, we 
	approximate the dynamics in 
	\eqref{eq:compact_dynamics} to a discrete-time linear system. Namely, for each reference vector 
	$\mathbf{x}_{ref}[k]$ (for $k = 1,...,n$, where $n$ is the prediction horizon length), we compute 
	the approximate linear, discrete system matrices 
	$\mathbf{A}_d[k]$ and $\mathbf{B}_d[k]$. 
	
	We first substitute the feet locations $\mathbf{p}_i$ obtained from the contact sequence task 
	into $\mathbf{B}(\Theta,\mathbf{p}_{LF},...,\mathbf{p}_{LH})$ for every 
	contact configuration 
	at time instant 
	$k$. However, $\mathbf{A}(\Theta)$ and 
	$\mathbf{B}(\Theta,\mathbf{p}_{LF},...,\mathbf{p}_{LH})$ are still dependent on the body orientation (in a nonlinear fashion).
%	still dependent in a nonlinear fashion on the body orientation. 
	To obtain the linear, discrete 
	time versions of these matrices, we follow a similar argument to 
	\cite{iros18dicarlo}. Assuming 
	that the \ac{MPC}-based controller will follow sufficiently close the reference trajectory 
	given by $\mathbf{x}_{ref}[k]$, we substitute the values of $\Theta_{ref}[k]$ into system matrices 
	$\mathbf{A}(\Theta)$ and $\mathbf{B}(\Theta,\mathbf{p}_{LF},...,\mathbf{p}_{LH})$. We also consider the values of $\theta_{ref}$ and $\phi_{ref}$ 
	computed by the \ac{COM} reference trajectory. We then discretize the system matrices using a 
	\ac{ZOH}. The discrete-time linear system dynamics can be described as
	\begin{equation}
		\mathbf{x}[k + 1] = \mathbf{A}_d[k] \mathbf{x}[k] + \mathbf{B}_d[k]\mathbf{u}[k]
		\label{eq:discrete_system}
	\end{equation}
	\paragraph{Model Predictive Control} \changerev{W}e can obtain a discrete 
	time evolution of the system by 
	successive substitution of states $\mathbf{x}[k]$ into \eqref{eq:discrete_system} to 
	obtain the 
	state evolution from $k = 0$ to $k = n$. Then, we can describe the dynamics as
	\begin{equation}
		\mathbf{X} = \bar{\mathbf{A}} \mathbf{x}_0 + \bar{\mathbf{B}}\bar{\mathbf{u}}
	\end{equation}
	where $\mathbf{X} \in \Rnum^{15n}$ is the stacked vector of states along 
	the prediction horizon $\mathbf{X} = \left[\mathbf{x}^\top[1], \text{ } 
	..., \text{ } \mathbf{x}^\top[n]\right]^\top$, $\bar{\mathbf{A}}\in 
	\Rnum^{15n\times 15n}$ and $\bar{\mathbf{B}}\in \Rnum^{15n \times 12n}$ are 
	the matrices built by successive substitution,
%	 along the prediction horizon, 
	$\mathbf{x}_0 \in \Rnum^{15}$ is the actual robot state vector and 
	$\bar{\mathbf{u}} \in \Rnum^{12n}$ is the stacked vector of ground reaction 
	forces
%	 along the prediction horizon 
	 $\bar{\mathbf{u}} = 
	\left[\mathbf{u}^\top[0], \text{ }, ..., \text{ } 
	\mathbf{u}^\top[n-1]\right]^\top$. 	
	We formulate the optimization problem to minimize the weighted least-squares error between the states and the 
	reference along the prediction horizon. We enforce the gait pattern 
	$\pazocal{G}$ and friction consistency by setting 
	appropriate 
	constraints. 
	\change{W}e solve 
%	the following problem
%	the 
%	optimization problem
%	
\begin{align}
	&\min_{\bar{\mathbf{u}}} &  \left\lVert \mathbf{X} - 
	\mathbf{X}_{ref}\right\rVert^2_\mathbf{L} + 
	\left\lVert\bar{\mathbf{u}}\right\rVert^2_{\mathbf{K}} &\nonumber\\
	&\text{subject to} & -\mathbf{\mu}\bar{\mathbf{u}}_{z} \le \bar{\mathbf{u}}_{x} \le \mathbf{\mu}\bar{\mathbf{u}}_{z} 
	&                  & -\mathbf{\mu}\bar{\mathbf{u}}_{z} \le \bar{\mathbf{u}}_{y} \le \mathbf{\mu}\bar{\mathbf{u}}_{z} &\nonumber\\
	&                  & \mathbf{u}_{min} \le \bar{\mathbf{u}}_{z}            \le \mathbf{u}_{max} & 
	&                  \mathrm{G}(\pazocal{G}) \bar{\mathbf{u}}                         = \mathbf{0} 
	\label{eq:optimization_problem}
\end{align} 
	where $\mathbf{X}_{ref} \in \Rnum^{15n}$ is the stacked vector of desired 
	states along the prediction horizon\footnote{We redefine
	$\mathbf{x}_{ref}[k] = [\Theta_{ref}^{\top}[k]\text{  } 
	\mathbf{r}_{ref}^{\top}[k]\text{  }  
	(\mathbf{T}(\Theta_{ref}[k])\dot{\Theta}_{ref}[k])\change{^\top}\text{ 
	 }\dot{\mathbf{r}}_{ref}^{\top}[k]\text{  }	
	\change{\mathbf{g}^\top}]^\top$ 
	to 	match the state definition of $\mathbf{x}$}, 
	vectors 
	$\bar{\mathbf{u}}_x\in\Rnum^{4n}$, $\bar{\mathbf{u}}_y\in\Rnum^{4n}$ and 
	$\bar{\mathbf{u}}_z\in\Rnum^{4n}$ correspond to the components of vector 
	$\bar{\mathbf{u}}$ associated to $x$, $y$ and $z$, 
	respectively, of the \acp{GRF}, $\mu\in \Rnum$ is the friction coefficient, 
%	between the ground and the feet, 
	$\mathbf{u}_{min}\in\Rnum^{4n}$ and 
	$\mathbf{u}_{max} \in \Rnum^{4n}$ are the limits on the $z$ component of 
	$\bar{\mathbf{u}}$, matrix $\mathrm{G} \in \Rnum^{12n\times12n}$ is a 
	matrix that selects the components of the \acp{GRF} that are in contact 
	according to gait $\pazocal{G}$, and matrices $\mathbf{L}$ and $\mathbf{K}$ 
	are weighting matrices. \changerev{This optimization problem}
%	The optimization problem defined by 
%	\eqref{eq:optimization_problem} 
	is a \ac{QP} and can be efficiently solved 
	by several off-the-shelf solvers. After solving 
	\changerev{\eqref{eq:optimization_problem}}, 
	\changerev{we compute the desired wrench coming from the \ac{MPC}} as
%	the problem in 
%	\eqref{eq:optimization_problem}, 
%	we take the first 12 entries of the 
%	optimized input vector $\bar{\mathbf{u}}\changerev{^*}$, 
%	which 
%	\changerev{are}
%	the set of \acp{GRF} at time instant $k = 1$ 
%	and compute the desired wrench 
%	coming from the \ac{MPC}-based controller as
%	
\change{
	\begin{align}
		\mathbf{w}_{MPC} = \sum_{i=1}^{l}\begin{bmatrix}
							\mathbf{p}_i \times \mathbf{F}^*_{i} \\
							\mathbf{F}^*_{i}
							\end{bmatrix}
	\end{align}
}where $\mathbf{F}^*_{i}$ is the optimized \ac{GRF} of foot $i$, 
	\changerev{extracted from the
	first 12 entries of the 
	optimized control input vector $\bar{\mathbf{u}}\changerev{^*}$.}
%	which 
%	\changerev{are}
%	the set of \acp{GRF} at time instant $k = 1$
%	
	\paragraph{Leg Inertia Compensation} \changerev{T}he \ac{MPC} model used 
	for prediction neglects leg 
	inertia. This assumption is acceptable for quasi-static motions. However, if the leg-body weight ratio is significantly large, the wrench exerted 
	by the legs on the body plays a significant role in the dynamics. We compensate for these 
	effects in a simple, yet effective, manner. 
%	We compute the influence of the joint acceleration on the body using the top-right cross-term of the joint-space inertia matrix. Specifically
	The floating-base dynamics of a robot can be described by
	\begin{equation}
		\begin{bmatrix}
			\mathbf{M}_u & \mathbf{M}_{ua} \\
			\mathbf{M}_{au}	 & \mathbf{M}_a
		\end{bmatrix}
		\begin{bmatrix}
			\dot{\mathbf{v}} \\
			\ddot{\mathbf{q}}_j
		\end{bmatrix}
		+
		\begin{bmatrix}
			\mathbf{h}_u \\
			\mathbf{h}_a
		\end{bmatrix} =
		\begin{bmatrix}
			\mathbf{0} \\
			\tau_j
		\end{bmatrix}
		+
		\begin{bmatrix}
			\mathbf{J}_{c,u}^\top \\
			\mathbf{J}_{c,a}^\top 
		\end{bmatrix}
		\mathbf{F}
		\label{eq:robot_dynamics}
	\end{equation}
	where $\mathbf{v}\in \Rnum^6$ is the floating-base robot velocity, $\mathbf{q}\in \Rnum^{n_j}$ 
	is the 
	joint 
	configuration, $\mathbf{M}_u \in \Rnum^{6\times 6}$ and $\mathbf{M}_a \in \Rnum^{6\times n_j
	}$ are the direct un-actuated and actuated parts of the joint-space inertia 
	matrix, whereas $\mathbf{M}_{ua} \in \Rnum^{6\times n_j}$ and 
	$\mathbf{M}_{au} \in \Rnum^{n_j\times 6}$ correspond to the cross terms 
	between actuated and un-actuated parts of the joint-space inertia matrix, 
	$\mathbf{h}_u \in 
	\Rnum^6$ 
	and $\mathbf{h}_a\in \Rnum^{n_j}$ are the un-actuated and actuated vectors of Coriolis, 
	centrifugal and 
	gravitational terms, $\tau_j \in \Rnum^{n_j}$ is the vector of joint torques, $\mathbf{J}_{c,u} 
	\in 
	\Rnum^{n_c\times 6}$ and $\mathbf{J}_{c,a}\in\Rnum^{n_c\times n_j}$ are the un-actuated and 
	actuated contact Jacobians  and $\mathbf{F}$ is the vector of \acp{GRF}. 
	The matrix $\mathbf{M}_{ua}$ maps the joint accelerations to the robot 
	spatial force acting on the floating-base of the robot, namely
	\begin{equation}
		\mathbf{w}_l = \mathbf{M}_{ua} \ddot{\mathbf{q}}_j
		\label{eq:ic_actual}
	\end{equation} 
	In \eqref{eq:ic_actual}, the $\mathbf{w}_l$ can be computed directly using 
	measurements coming from the sensors. 
	However, using the actual joint 
	acceleration might lead to high frequency wrench signals. Instead, 
	We 
	use the desired joint acceleration $\ddot{\mathbf{q}}_{j,d}$ coming from 
	the torque mapper (see Fig. 
	\ref{fig:strategy}). Then, the leg inertia compensation wrench is given by 
	$\mathbf{w}_l =  \mathbf{M}_{ua} \ddot{\mathbf{q}}_{j,d}$. Thus, the total 
	desired wrench is given by $\mathbf{w}_{d} = \mathbf{w}_{MPC} + 
	\mathbf{w}_l$\changerev{\footnote{\changerev{This total wrench is 
	distributed between the contact feet using the torque mapper shown in Fig. 
	\ref{fig:strategy}~\cite{focchi2016}}}}.

	\section{Results}
	\label{section:results}
	
		\label{section:implementation}
		\begin{figure*}[t]
		\centering
		\begin{subfigure}[t]{0.6\linewidth}
			\centering
			\includegraphics[width=1\linewidth]{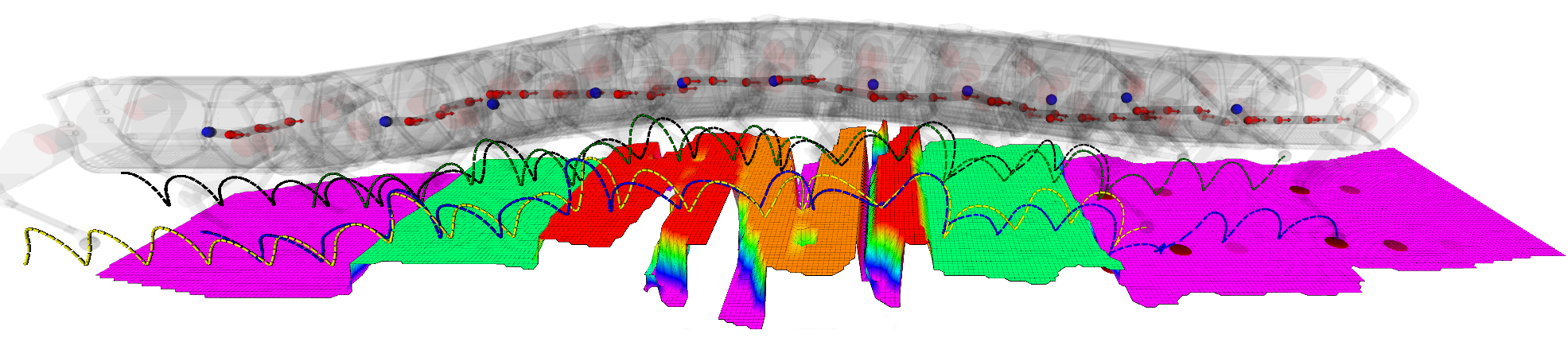}
		\end{subfigure}
		\begin{subfigure}[t]{0.35\linewidth}
			\centering
			\includegraphics[width=1\linewidth]{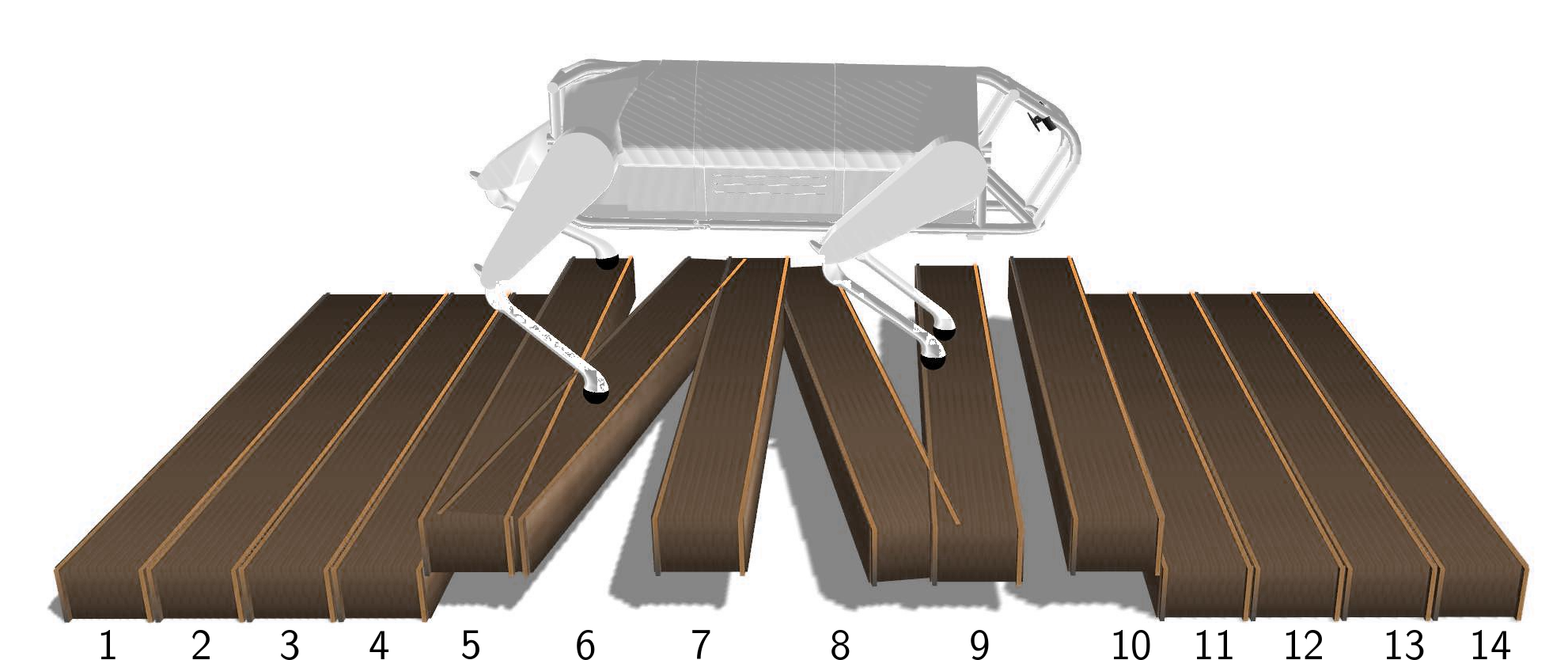}
		\end{subfigure}		
		\caption{\small \changerev{L}eft: series of snapshots of the HyQReal 
		robot moving through the scenario. Blue spheres correspond to the 
		position of the center of mass at the moment when the snapshot was 
		taken and the red spheres show the reference position for the \ac{COM} 
		along the prediction horizon. The positions of the feet are indicated 
		by the colored dashed lines. The elevation map is built using the 
		vision sensors. Right: scenario designed to test the locomotion 
		strategy proposed in this paper. Each beam is \SI{15}{\centi \meter} 
		height and \SI{20}{\centi \meter} wide. Beams 1 to 4 and 11 to 14 are 
		located at ground level. Beams 5 to 7 and 10 are located \SI{15}{\centi 
		\meter} above ground level. Beam 8 is located \SI{12}{\centi \meter} 
		above ground.}
		\label{fig:performance_task_sim}
		\vspace{-0.275cm}
	\end{figure*}
%	In this section we detail the implementation setup and the obtained simulation results showing, with respect to our previous work [16], the improvements in foothold prediction, locomotion safety and reliability, and in the robot capabillity to traverse more difficult terrain.
%	the simulation results obtained with respect to improvement in foothold prediction, safety and reliability in locomotion and the capability to traverse more difficult terrain with respect to our previous work \cite{villarreal19ral}.	
%	\subsection{Implementation}
%	\label{section:implementation}
%\vspace{-0.2cm}
	We performed simulations on HyQReal \cite{hyqreal19irim}, a hydraulically 
	actuated 
	quadruped robot. 
	The leg configuration of the robot is shown in 
	Fig.~\ref{fig:strategy}. 
%	It weighs \SI{130}{\kilogram} with the on-board \ac{HPUs} and battery, and 
%	approximately \SI{105}{\kilogram} when external electric/hydraulic power is used. It is 
%	\SI{1.3}{\meter} long and \SI{0.9}{\meter} tall. 
	We use Gazebo \cite{gazebo} to perform our simulations. Control commands are executed at a frequency of \SI{250}{\hertz}. 
	Wrench values from the \ac{MPC} controller $\mathbf{w}_{MPC}$ are 
	sent at a maximum frequency of \SI{25}{\hertz} and we use a \ac{ZOH} in 
	between control signals. The prediction horizon is set to comprise 2 gait 
	cycles, partitioned in 20 samples. 
	We solve the \ac{QP} in \eqref{eq:optimization_problem} with a modified 
	version of uQuadProg++ \cite{quadprog++} to work with the C++ linear 
	algebra library, Eigen. \change{The signal $\mathbf{w}_{MPC}$ is computed 
	at \SI{25}{\hertz} to give enough time to the solver to compute a feasible 
	solution. }The leg inertia compensation wrench $\mathbf{w}_l$ is computed 
	at task frequency. 
	The mapping is done using the Grid Map interface from \cite{fankhauser14clawar}. 
	Weighting marices are chosen as $\mathbf{L} = \mathbf{1}_{15n}$ and 
	$\mathbf{K}=(1\times10^{-9})\mathbf{1}_{12n}$, where $\mathbf{1}_{a}$ 
	defines the $a\times a$ identity matrix. 
	\change{\subsection*{Simulations}}
		\begin{figure}[h!]
		\centering
		\includegraphics[width=0.98\linewidth]{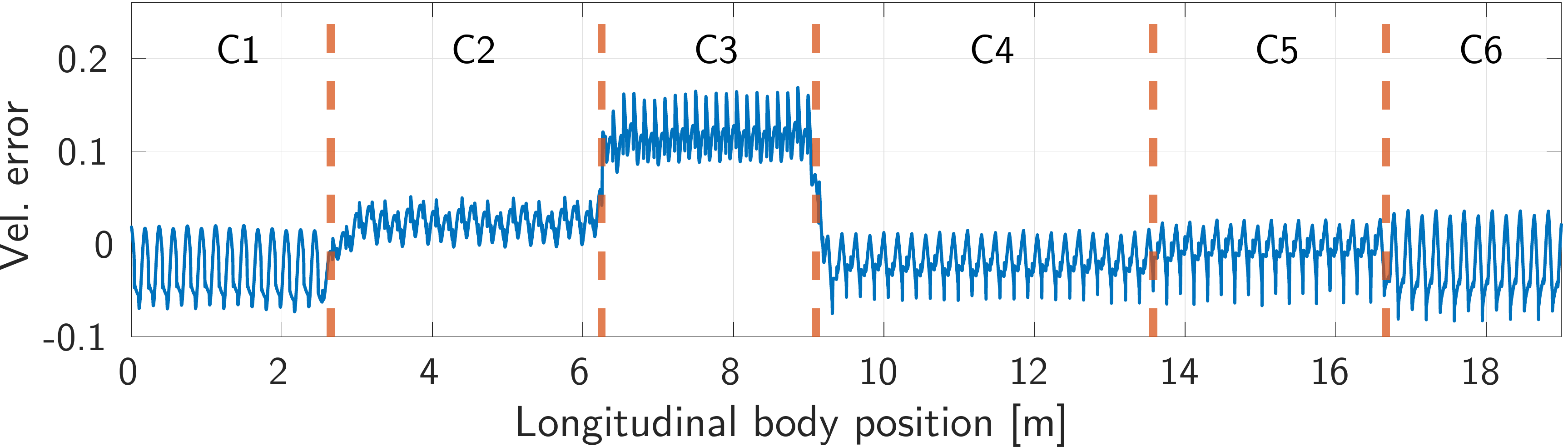}
		\caption{\small \changerev{V}elocity with different control 
		strategies. The red dashed lines indicate the moments when the
		configuration was changed. The controller configurations are: C1 = QP + 
		LI + GC; C2 = QP + LI + GC + IC; C3 = QP + LI + IC; C4 = MPC + LI + IC; 
		C\change{5} = MPC + IC; C\change{6} = MPC. The abbreviations stand for: 
		a) \ac{QP}: standard \changerev{\underline{\ac{QP}}} trunk controller, 
		b) LI: stance \changerev{\underline{l}}eg 
		\change{joint} \changerev{\underline{i}}mpedance, c) GC: 
		\changerev{\underline{g}}ravity \changerev{\underline{c}}ompensation, 
		d) IC: leg 
		\changerev{\underline{i}}nertia \changerev{\underline{c}}ompensation 
		and, e) \ac{MPC}: \changerev{\underline{m}}odel 
		\changerev{\underline{p}}redictive \changerev{\underline{c}}ontroller.}
		\label{fig:velocity_test}
		\vspace{-0.3cm}
	\end{figure}
	\label{section:simulation_results}
		\bgroup
	\def\arraystretch{1.0}
	\setlength{\tabcolsep}{4pt}
	\begin{table}[t]
		\centering
		\caption{Root mean square and maximum absolute value of the foothold prediction 
			error}
		\label{table:max_error}
		\begin{tabular}{c|c|cccc}
			&            & LF     & RF     & LH     & RH     \\ \hline
			\multirow{2}{*}{QP+LI+GC}  & $RMS(e)$   & 0.012  & 0.012  & 0.012  & 0.012  \\ \cline{2-6} 
			& $\max |e|$ & 0.095  & 0.095  & 0.088  & 0.082  \\ \hline
			\multirow{2}{*}{MPC+IC} & $RMS(e)$   & 0.009  & 0.007  & 0.007  & 0.009  \\ \cline{2-6} 
			& $\max |e|$ & 0.0740 & 0.0611 & 0.0604 & 0.0761
		\end{tabular}
	\vspace{-0.3cm}
	\end{table}
	\egroup
	\begin{figure}
		\centering		
		\includegraphics[width=0.95\linewidth]{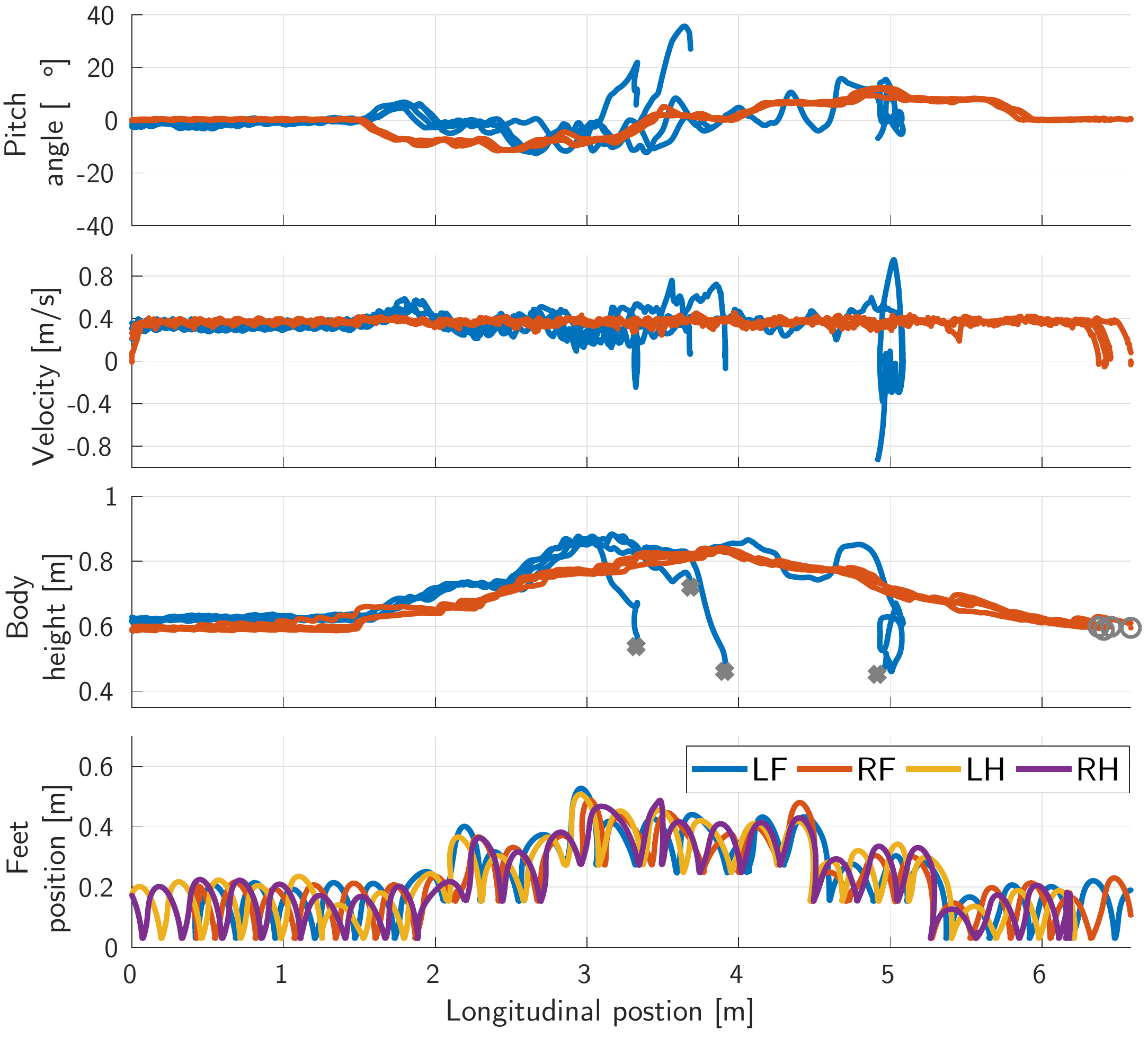}
		\caption{\small \changerev{R}esults for the scenario crossing 
		simulation. The top 
		three plots show pitch angle, velocity error\changerev{,} and body 
		height. Blue 
		lines correspond to trials with configuration \change{C1} and red lines 
		correspond to 
		\change{C5}. Foot trajectories corresponding to one 
		of the succesful trials are shown at the bottom part of the figure.}
		\label{fig:performance_task}
		\vspace{-0.3cm}
	\end{figure}	
	We perform three different simulations to assess the improvements in foothold prediction and locomotion robustness. Below we explain in detail the outcome of these tests.
	\paragraph{Leg Inertia Compensation}
	\changerev{W}e perform simulations commanding the robot to trot on flat 
	terrain with a forward velocity $\mathbf{V}_f$ of \changerev{magnitude} 
	\SI{0.5}{\meter / 
	\second}, a step frequency $f_s$ of \SI{1.4}{\hertz}, and a duty factor 
	$D_f$ of 0.6. We start the simulation with our previous trunk controller 
	\cite{focchi2016}. We keep $\mathbf{V}_f$ and change the controller 
	configuration as the robot continues to trot. There are six possible 
	configurations shown in Fig.~\ref{fig:velocity_test}\changerev{,} which 
	combine the 
	following control components: 
	\begin{inlineenum}
		\item \ac{QP}: standard \changerev{\underline{\ac{QP}}} trunk controller
		\item  LI: stance \changerev{\underline{l}}eg \change{joint} 
		\changerev{\underline{i}}mpedance 
		\change{(PD 
		controller 
		at joint level)}
		\item  GC: \changerev{\underline{g}}ravity 
		\changerev{\underline{c}}ompensation
		\item  IC: leg \changerev{\underline{i}}nertia 
		\changerev{\underline{c}}ompensation and
		\item  \ac{MPC}: \changerev{\underline{m}}odel 
		\changerev{\underline{p}}redictive \changerev{\underline{c}}ontroller.
	\end{inlineenum}
	Figure~\ref{fig:velocity_test} shows the error in velocity with respect to 
	the commanded $\mathbf{V}_f$ for one of the trials. The vertical dashed red 
	lines indicate the moments when the controller configuration was changed. 
	We check six different configurations, although we would like to stress 
	that configuration C3 \changerev{(see Fig.~\ref{fig:velocity_test})} acts 
	merely as a a transition between the standard QP 
	and the MPC. This is because the MPC controller already compensates for the 
	gravitational effects in the model. It can be noticed that when the inertia 
	compensation wrench is applied, the accelerations of the body are greatly 
	reduced. The best performing configuration corresponds to the configuration 
	C5 (fifth portion of graph in Fig.~\ref{fig:velocity_test}). Under this 
	configuration, the robot dynamics resemble more those of the \ac{MPC} model 
	(which neglects leg inertia), since the leg inertia is being accounted for 
	outside of the optimization. 
	\change{C2 presents larger errors in 
	velocity tracking, but its response is smoother with respect to C5. The 
	sharp changes in velocity in C5 could be reduced by carefully choosing the 
	weights in  $\mathbf{K}$ and  $\mathbf{L}$ of 
	\eqref{eq:optimization_problem}.
%	are related to the coarse tunning performed for the $\mathbf{K}$ 
%	matrix in \eqref{eq:optimization_problem}. 
%	This tunning should be done when carrying out experiments.
	}
	\paragraph{Foothold Predictions and Robustness in the Presence of Disturbances}
	\changerev{F}or the second simulation\changerev{,} the robot is also 
	commanded to trot 
	on flat 
	terrain with the same gait parameters as in the first simulation. This 
	time\changerev{,} 
	we perturb \changerev{the robot} three times with \SI{700}{\newton} of 
	force 
	with a duration 
	of \SI{0.1}{\second}. Table \ref{table:max_error} shows the \ac{RMS} error 
	and the maximum absolute value of the foothold prediction error for this 
	simulation. The table helps us to compare the previous controller 
	configuration with the \ac{MPC}-based controller with leg inertia 
	compensation. The \ac{RMS} \changerev{prediction} error when using \ac{MPC} 
	and leg inertia 
	compensation is between 25\% and 41\% less with respect to the previous 
	controller configuration. This represents \change{between} \SI{3}{\milli 
	\meter} \change{and} 
	\SI{5}{\milli \meter} of improvement. However, even if the average of the 
	error is low in both cases, a single wrong prediction compromises the robot 
	stability. \change{T}he maximum absolute value of the error is more 
	representative of the reliability of the prediction under disturbances. In 
	this case, the reduction of the error is between 7\% and 36\%. This 
	represents between \SI{0.6}{\centi \meter} and \SI{3}{\centi \meter} of 
	reduction of the foothold prediction error when using the \ac{MPC} with the 
	leg inertia compensation. 
		
	\paragraph{Locomotion on Challenging Terrain}	
	\changerev{T}o verify the improvement in performance regarding locomotion 
	on difficult 
	terrain, we designed the 
%	challenging 
	scenario in 
	Fig.~\ref{fig:performance_task_sim}. The robot is commanded to trot with a 
	forward velocity of \SI{0.4}{\meter / \second}, a step frequency of 
	\SI{1.4}{\hertz} and a duty factor of 0.6. \changerev{W}e use the \ac{VFA} 
	\changerev{to select appropriate footholds} with two different control 
	configurations 
	\begin{inlineenum}
		\item QP + LI + GC \changerev{(C1)} and
		\item \ac{MPC} + IC \changerev{(C5)}.
	\end{inlineenum}
	To test the performance repeatability we did four trials with each 
	configuration. Figure~\ref{fig:performance_task} \changerev{shows the}
%	contains the plots 
%	corresponding to 
	pitch angle, forward velocity, body height and an example 
	of the foot trajectories for one of the trials with 
	\changerev{configuration C5}.  
	Figure \ref{fig:performance_task_sim} shows 11 overlapped snapshots of the RVIZ visualization as the robot crosses the scenario, builds the map, and adjusts its footholds on the fly.  
	Figure \ref{fig:performance_task_sim} also shows the reference trajectory of the center of mass given at the specific moment when the snapshot was taken, and the foot trajectories as the robot moves through the scenario.
	
	\change{Figure~\ref{fig:performance_task} shows} that all four different 
	trials using \changerev{configuration C5} were successful and the 
	variations 
	in linear velocity, pitch and body height are significantly reduced with 
	respect to \changerev{C1}. For this last configuration, the 
	robot was not able to reach the end of the scenario in any of the 
	trials\change{, mainly due to errors in foothold prediction and variations 
	on the body velocity}. 
	This task shows the mutual benefits between the \ac{VFA} and the 
	\ac{MPC}. The foothold prediction error is reduced when 
	using the strategy here presented. Specifically, in the case of the 
	\ac{MPC} + IC, for all four trials and all legs, the 
	maximum absolute value of the error in foothold prediction was 
	\SI{10}{\centi\meter}, while in the case of the previous configuration the 
	error was \SI{14}{\centi\meter}. 

	\section{Conclusions and future work}
	\label{section:conclusions}
	 
	We developed a dynamic locomotion strategy to traverse difficult terrain 
	using visual information only coming from on-board sensors. We based this 
	strategy on the combination of an \ac{MPC}-based controller and a 
	\ac{CNN}-based foothold adaptation \changerev{scheme} (namely the 
	\ac{VFA}). We showed that the interaction between these approaches is 
	mutually beneficial and improves locomotion reliability and robustness. We 
	also demonstrated that considering a compensation term accounting for the 
	wrench due to the inertia of the legs improves the performance of the 
	\ac{MPC}-based controller, due to a closer resemblance to the model used 
	for state prediction.
	The various simulations validated these improvements. \change{A major 
	limitation is related to the maximum frequency at which the \ac{MPC} 
	controller can be computed. For the current prediction horizon length, this 
	is limited to \SI{25}{\hertz}.}
	As future work\changerev{,} we plan to validate the strategy developed 
	\changerev{here}
	in experiments with the hydraulically actuated quadruped robot HyQReal. 	

	\bibliographystyle{IEEEtran}
	\bibliography{references}	
\end{document}